\documentclass[runningheads]{llncs}
\usepackage[T1]{fontenc}
\usepackage{graphicx}
\usepackage{times}
\usepackage{multicol}
\usepackage[bookmarks=true]{hyperref}

\usepackage{graphicx}
\usepackage{float}
\usepackage{rotating}
\usepackage{subcaption}
\usepackage{psfrag}
\usepackage{amsmath,amssymb,amsfonts,bbm,nicefrac,mathtools,lipsum}
\usepackage{empheq}
\usepackage{multirow}
\usepackage{latexsym}
\usepackage{caption}
\usepackage{balance}
\usepackage{colortbl}
\usepackage[table]{xcolor}
\usepackage[super]{nth}
\usepackage{fancyhdr}
\usepackage{epstopdf}
\usepackage{float}
\usepackage{booktabs,tabularx}
\usepackage{makecell}
\usepackage{dirtytalk}
\usepackage[sort&compress,numbers]{natbib}
\usepackage[noend]{algpseudocode}
\usepackage{siunitx}
\usepackage{academicons}
\usepackage{cancel}
\usepackage{xparse}
\makeatletter
\usepackage{xspace}
\DeclareRobustCommand\onedot{\futurelet\@let@token\@onedot}
\def\@onedot{\ifx\@let@token.\else.\null\fi\xspace}
\def\eg{e.g\onedot} 
\def\ie{i.e\onedot} 
\def\cf{cf\onedot}

\def\NAT@spacechar{~}%
\makeatother
\usepackage[capitalize]{cleveref}
\usepackage{soul}
\usepackage[ruled]{algorithm2e}
\SetAlCapSkip{10em}
\SetAlgoSkip{bigskip}

\usepackage[acronyms, shortcuts]{glossaries}
\glsdisablehyper

\newacronym{em}{EM}{expectation maximization}
\newacronym{ppo}{PPO}{proximal policy optimization}
\newacronym{drl}{DRL}{deep reinforcement learning}
\newacronym{il}{IL}{imitation learning}
\newacronym{ece}{ECE}{entropic cost equilibrium}
\newacronym{ioc}{IOC}{inverse optimal control}
\newacronym{lqr}{LQR}{linear-quadratic regulator}
\newacronym{lqg}{LQG}{linear-quadratic-Gaussian}
\newacronym{ilqg}{ilqg}{iterative linear-quadratic Gaussian}
\newacronym{ilqr}{iLQR}{iterative linear-quadratic regulator}
\newacronym{kkt}{KKT}{Karush–Kuhn–Tucker}
\newacronym{irl}{IRL}{inverse reinforcement learning}
\newacronym{mle}{MLE}{maximum likelihood estimation}
\newacronym{map}{MAP}{maximum a posteriori}
\newacronym{sem}{SEM}{standard error of the mean}
\newacronym{mpgp}{MPGP}{model-predictive game-play}
\newacronym[longplural=open-loop Nash equilibria,\glsshortpluralkey={OLNE}]{olne}{OLNE}{open-loop Nash equilibrium}
\newacronym{licq}{LICQ}{linear independence constraint qualification}
\newacronym{mpc}{MPC}{model-predictive control}
\newacronym[longplural=mathematical programs with equilibrium constraints,plural={MPEC}]{mpec}{MPEC}{mathematical program with equilibrium constraints}
\newacronym[longplural={partially observable Markov decision processes}]{pomdp}{POMDP}{partially observable Markov decision process}
\newacronym{posg}{POSG}{partially observable stochastic game}
\newacronym{nep}{NEP}{Nash equilibrium problem}
\newacronym{gnep}{GNEP}{generalized Nash equilibrium problem}
\newacronym[longplural={generalized Nash equilibria},\glsshortpluralkey={GNE}]{gne}{GNE}{generalized Nash equilibrium}
\newacronym{svo}{SVO}{social value orientation}
\newacronym{ukf}{UKF}{unscented Kalman filter}
\newacronym{ibr}{IBR}{iterated best response}
\newacronym{awgn}{AWGN}{additive white Gaussian noise}
\newacronym{iqr}{IQR}{interquartile range}
\newacronym{mcp}{MCP}{mixed complementarity problem}
\newacronym{ift}{IFT}{implicit function theorem}
\newacronym{nn}{NN}{neural network}
\newacronym{rrt}{RRT}{rapidly exploring random trees}
\newacronym{idm}{IDM}{intelligent driver model}
\newacronym{lstm}{LSTM}{long short-term
memory}
\newacronym{lq}{LQ}{linear-quadratic}
\newacronym{gail}{GAIL}{generative adversarial imitation learning}
\newacronym{dagger}{DAgger}{dataset aggregation}
\newacronym{vae}{VAE}{variational autoencoder}
\newacronym{vi}{VI}{variational inference}
\newacronym{cgan}{(C)GAN}{(conditional) generative adversarial network}
\newacronym{cvae}{(C)VAE}{(conditional) variational autoencoder}
\newacronym{rnn}{RNN}{recurrent neural network}
\newacronym{fov}{FOV}{field-of-view}
\newacronym{ros}{ROS}{Robot Operating System}
\newacronym{kl}{KL}{Kullback–Leibler}
\newacronym{elbo}{ELBO}{evidence lower bound}
\newacronym{sgd}{SGD}{stochastic gradient descent}
\newacronym{birl}{BIRL}{Bayesian inverse reinforcement learning}
\newacronym{sqp}{SQP}{sequential quadratic programming}
\newacronym{ood}{OOD}{out-of-distribution}
\newacronym{stl}{STL}{signal temporal logic}
\newcommand{\observation}{y}
\newcommand{\dkl}[2]{D_\mathrm{KL}\left(#1 \parallel #2\right)}

\newcommand{\gameParams}{\theta}
\newcommand{\encoderParams}{\psi}
\newcommand{\decoderParams}{\phi}
\newcommand{\encoder}[1][\encoderParams]{e_{#1}}
\newcommand{\decoder}[1][\decoderParams]{d_{#1}}
\DeclareRobustCommand{\expectedValue}[2]{\mathbb{E}_{#1}\left[#2\right]}
\newcommand{\gradcondition}[1]{\textbf{C#1}}

\newcommand{\dataset}{\mathcal{D}}
\newcommand{\latent}{z}
\newcommand{\traj}{\tau}
\newcommand{\solveGame}{\mathcal{T}_{\game}}

\newcommand{\noise}{\epsilon}
\newcommand{\remap}{r}
\newcommand{\loss}{\mathcal{L}}
\newcommand{\losssample}{\ell}
\newcommand{\normaldist}{\mathcal{N}}
\newcommand{\transpose}{\top}
\newacronym{iid}{iid}{independent and identically distributed}
\newacronym{ji}{JI}{Jensen's inequality}
\newacronym{pdf}{PDF}{probability density function}

\newacronym{hj}{HJ}{Hamilton–Jacobi}

\newcommand{\given}{\mid}

\newcommand{\cost}{J}

\newcommand{\costparams}{\theta}

\newcommand{\game}{\Gamma}

\newcommand{\reals}{\mathbb{R}}

\newcommand{\revised}[1]{{\leavevmode\color{black}#1}}
\newcommand{\revisedInline}[1]{\textcolor{black}{#1}}
\newcommand{\draft}[1]{{\leavevmode\color{magenta}#1}}

\graphicspath{{./Images/}}
\begin{document}
\title{Auto-Encoding Bayesian Inverse Games}
\author{Xinjie Liu\inst{1}$^*$\orcidID{0000-0001-7826-4506} \and
Lasse Peters\inst{2}$^*$\orcidID{0000-0001-9008-7127} \and
Javier Alonso-Mora\inst{2}\orcidID{0000-0003-0058-570X} \and
Ufuk Topcu\inst{1}\orcidID{0000-0003-0819-9985} \and
David Fridovich-Keil\inst{1}\orcidID{0000-0002-5866-6441}
}
\authorrunning{X. Liu et al.}
\institute{The University of Texas at Austin, Austin, TX 78712, USA
\email{\{xinjie-liu,utopcu,dfk\}@utexas.edu}
\and
Delft University of Technology, Delft, 2600 AA, Netherlands
\email{\{l.peters,j.alonsomora\}@tudelft.nl}\\
}

\makeatletter 
\def\thanks#1{\protected@xdef\@thanks{\@thanks \protect\footnotetext{#1}}}
\thanks{$^{*}$Equal contribution}
\makeatother

\maketitle              %
\begin{abstract}

When multiple agents interact in a common environment, each agent's actions impact others' future decisions, and noncooperative dynamic games naturally capture this coupling.
In interactive motion planning, however, agents typically do not have access to a complete model of the game, \eg, due to unknown objectives of other players.
Therefore, we consider the \emph{inverse} game problem, in which some properties of the game are unknown a priori and must be inferred from observations.
Existing \ac{mle} approaches to solve inverse games provide only point estimates of unknown parameters without quantifying uncertainty, and perform poorly when many parameter values explain the observed behavior.
To address these limitations, we take a Bayesian perspective and construct \emph{posterior distributions} of game parameters.
To render inference tractable, we employ a \ac{vae} with an embedded differentiable game solver.
This structured \ac{vae} can be trained from an unlabeled dataset of observed interactions, naturally handles continuous, multi-modal distributions, and supports efficient sampling from the inferred posteriors without computing game solutions at runtime. 
Extensive evaluations in simulated driving scenarios demonstrate that the proposed approach successfully learns the prior and posterior game parameter distributions, provides more accurate objective estimates than \ac{mle} baselines, and facilitates safer and more efficient game-theoretic motion planning.

\end{abstract}

\section{Introduction}
\label{sec:intro}

\begin{figure}[t]
  \centering
  \includegraphics[width=0.5\linewidth]{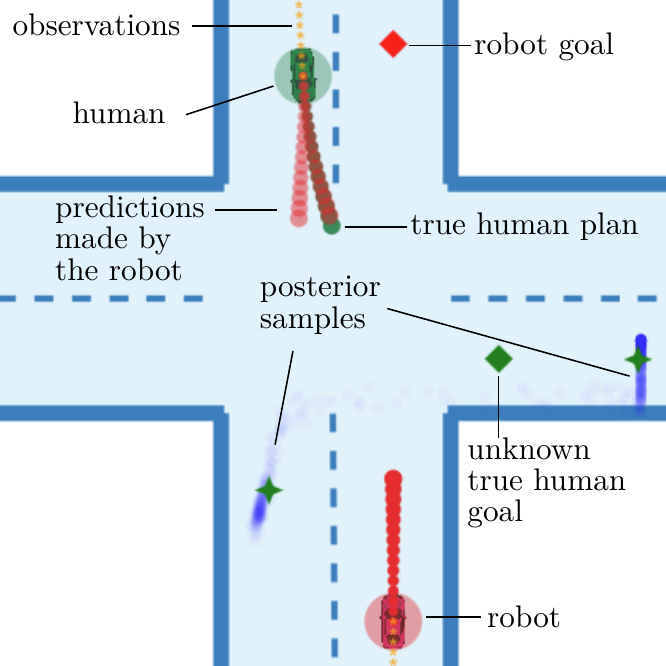}
  \caption{A robot interacting with a human driver whose goal position is unknown.
   We embed a differentiable game solver in a structured variational autoencoder to infer the \emph{distribution} of the human's objectives based on observations of their behavior.}
  \label{fig:motivation}
\end{figure}

Autonomous robots often need to interact with other agents to operate seamlessly in real-world environments.
For example, in the scenario depicted in~\cref{fig:motivation}, a robot encounters a human driver while navigating an intersection.
In such settings, coupling effects between agents significantly complicate decision-making: if the human acts assertively and drives straight toward its goal, the robot will be forced to brake to avoid a collision.
Hence, the agents compete to optimize their individual objectives.
\textit{Dynamic noncooperative game theory}~\cite{Basar1998games} provides an expressive framework to model such interaction among rational, self-interested agents.

In many scenarios, however, a robot must handle interactions under incomplete information, \eg, without knowing the goal position of the human driver in~\cref{fig:motivation}.
To this end, a recent line of work~\cite{waugh2013computational,kopf2017inverse,inga2019inverse,awasthi2020inverse,ROTHFU201714909,qiu2023identifying,liu2023learning,peters2023ijrr,li2023cost,mehr2023maxent} solves \textit{inverse dynamic games} to infer the unknown parameters of a game---such as other agents' objectives---from observed interactions.

A common approach to solve inverse dynamic games uses \acf{mle} to find the most likely parameters given observed behavior~\cite{liu2023learning,peters2023ijrr,li2023cost,mehr2023maxent,qiu2023identifying,kopf2017inverse,ROTHFU201714909,inga2019inverse,awasthi2020inverse}.
However, \ac{mle} solutions provide only a point estimate without any uncertainty quantification and can perform poorly in scenarios where many parameter values explain the observations~\cite{li2023cost}, yielding overconfident, unsafe motion plans~\cite{peters2024ral}.

Bayesian formulations of inverse games mitigate these limitations of \ac{mle} methods by inferring a posterior \emph{distribution}, \ie, belief, over the unknown parameters~\cite{peters2020accommodating,le2021lucidgames}.
Knowledge of this distribution allows the robot to account for uncertainty and generate safer yet efficient plans~\cite{le2021lucidgames,schwarting2021stochastic,hu2022active,peters2024ral}.

Unfortunately, exact Bayesian inference is typically intractable in dynamic games, especially when dynamics are nonlinear.
Prior work~\cite{le2021lucidgames} alleviates this challenge by using an \ac{ukf} for approximate Bayesian inference.
However, that approach is limited to unimodal uncertainty models and demands solving multiple games per belief update, thereby posing a computational challenge.

The main contribution of this work is a framework for tractable Bayesian inference of posterior distributions over unknown parameters in dynamic games\footnote{\href{https://xinjie-liu.github.io/projects/bayesian-inverse-games}{Supplementary video: \color{magenta}https://xinjie-liu.github.io/projects/bayesian-inverse-games}}. 
To this end, we approximate exact Bayesian inference with a structured \acf{vae}.
\revised{Unlike conventional \acp{vae}, our method embeds a differentiable game solver in the decoder, thereby facilitating unsupervised learning of an interpretable behavior model from an unlabeled dataset of observed interactions.}
At runtime, the proposed approach can generate samples from the predicted posterior without solving additional games.
As a result, our approach naturally captures continuous, multi-modal beliefs, and addresses the limitations of \ac{mle} inverse games without resorting to the simplifications or computational complexity of existing Bayesian methods.
\draft{
}

Through extensive evaluations of our method in several simulated driving scenarios, we support the following key claims.
The proposed framework (\romannumeral 1) learns the underlying prior game parameter distribution from unlabeled interactions, and (\romannumeral 2) captures potential multi-modality of game parameters.
Our approach (\romannumeral 3) is uncertainty-aware, predicting narrow unimodal beliefs when observations clearly reveal the intentions of other agents, and beliefs that are closer to the learned prior in case of uninformative observations.
The proposed framework (\romannumeral 4) provides more accurate inference performance than \ac{mle} inverse games by effectively leveraging the learned prior information, especially in settings where multiple parameter values explain the observed behavior.
As a result, our approach (\romannumeral 5) enables safer downstream robot plans than \ac{mle} methods.
We also qualitatively demonstrate the scalability of our inference scheme in a 4-player intersection driving scenario.

\section{Related Work}
\label{sec:related}

This section provides an overview of the literature on dynamic game theory, focusing on both forward games (\cref{related-works-forward-games}) and inverse games (\cref{related-works-inverse-games}).

\subsection{Forward Dynamic Games}
\label{related-works-forward-games}
This work focuses on noncooperative games where agents have partially conflicting but not completely adversarial goals and make sequential decisions over time~\cite{Basar1998games}.
Since we assume that agents take actions simultaneously without leader-follower hierarchy and we consider coupling between agents' decisions through both objectives and constraints, our focus is on \acp{gnep}.
\Acp{gnep} are challenging coupled mathematical optimization problems.
Due to the computational challenges involved in solving such problems under feedback information structure~\cite{laine2021computation}, most works aim to find \acp{olne}~\cite{Basar1998games} instead, where players choose their action sequence---an open-loop strategy---at once. %
Open-loop \acp{gne} have been solved using iterated best response~\cite{williams2018best,spica2020real,wang2019game,spica2020real,wang2021tro,schwarting2019social}, sequential quadratic approximations~\cite{cleac2022algames,zhu2023sequential}, or by leveraging the \ac{mcp} structure of their first-order necessary conditions~\cite{mcp_ref,liu2023learning,peters2024ral}.
This work builds upon the latter approach.

\subsection{Inverse Dynamic Games}
\label{related-works-inverse-games}
Inverse games study the problem of inferring unknown game parameters, e.g. of objective functions, from observations of agents' behavior~\cite{waugh2013computational}.
In recent years, several approaches have extended single-agent \ac{ioc} and \ac{irl} techniques to multi-agent interactive settings.
Early approaches~\cite{ROTHFU201714909, awasthi2020inverse} minimize the residual of agents' first-order necessary conditions, given \emph{full} state-control observations, in order to infer unknown objective parameters.
This approach is further extended to maximum-entropy settings in~\cite{inga2019inverse}.

More recent work~\cite{peters2023ijrr} proposes to maximize observation likelihood while enforcing the \ac{kkt} conditions of \acp{olne} as constraints.
This approach only requires partial-state observations and can cope with noise-corrupted data.
Approaches~\cite{li2023cost,liu2023learning} propose an extension of the \ac{mle} approach~\cite{peters2023ijrr} to inverse feedback and open-loop games with inequality constraints via differentiable forward game solvers.
\revised{To amortize the computation of the \ac{mle},  works \cite{geiger2021learning,liu2023learning} demonstrate integration with \ac{nn} components.}

In general, \ac{mle} solutions can be understood as point estimates of Bayesian posteriors, assuming a \textit{uniform} prior~\cite[Ch.4]{pml1Book}.
When multiple parameter values explain the observations equally well, this simplifying assumption can result in ill-posed problems---causing \ac{mle} inverse games to recover potentially inaccurate estimates~\cite{li2023cost}. %
Moreover, in the context of motion planning, the use of point estimates without awareness of uncertainty can result in unsafe plans~\cite{peters2024ral,hu2022active}.

To address these issues, several works take a Bayesian view on inverse games~\cite{le2021lucidgames,peters2020accommodating}, aiming to infer a posterior \emph{distribution} while factoring in prior knowledge.
Since exact Bayesian inference is intractable in these problems, the belief update may be approximated via a particle filter \cite{peters2020accommodating}.
However, this approach requires solving a large number of equilibrium problems online to maintain the belief distribution, posing a significant computational burden.
A sigma-point approximation~\cite{le2021lucidgames} reduces the number of required samples but limits the estimator to unimodal uncertainty models.

\revised{
To obtain multi-hypothesis predictions tractably, works in~\cite{diehl2023energy} and \cite{lidard2023nashformer} integrate game-theoretic layers in \acp{nn} for motion forecasting.
Both methods show that the inductive bias of games improves performance on real-world human datasets.
However, both approaches are limited to the prediction of a fixed number of intents
and offer no clear Bayesian interpretation of the learned model.
}

To overcome the limitations of  \ac{mle} approaches while avoiding the intractability of exact Bayesian inference \revised{over continuous game parameter distributions}, we propose to approximate the posterior via a \ac{vae}~\cite{kingma2013auto} that embeds a differentiable game solver~\cite{liu2023learning} during training.
The proposed approach can be trained from an unlabeled dataset of observed interactions, naturally handles continuous, multi-modal distributions, and does not require computation of game solutions at runtime to sample from the posterior.

\section{Preliminaries: Generalized Nash Game}\label{sec:preliminaries}
In this work, we consider strategic interactions of self-interested rational agents in the framework of \emph{\acfp{gnep}}.
In this framework, each of $N$  agents seeks to unilaterally minimize their respective cost while being conscious of the fact that their \say{opponents}, too, act in their own best interest.
We define a parametric $N$-player \ac{gnep} as $N$ coupled, constrained optimization problems,
\begin{subequations}
\label{eq:forward-gnep}
\begin{align}
\mathcal{S}^i_\gameParams(\traj^{\neg i}) :=
\arg\min_{\traj_i}\quad  & J^i_\gameParams(\traj^i, \traj^{\neg i})\\
\mathrm{s.t.}
\quad & g^i_\gameParams(\traj^i, \traj^{\neg i}) \geq 0,
\end{align}
\end{subequations}
where $\gameParams \in \reals^p$ denotes a parameter vector whose role will become clear below,
and player $i\in[N]:=\{1,\ldots,N\}$ has cost function $J^i_\gameParams$ and private constraints $g^i_\gameParams$. %
Furthermore, observe that the costs and constraints of player~$i$ depend not only on their own strategy $\traj^i \in \reals^{m_i}$, but also on the strategy of all other players, $\traj^{\neg i} \in \reals^{\sum_{j\in[N]\setminus\{i\}}{m_j}}$.

\smallskip\noindent\textbf{Generalized Nash Equilibria.}
For a given parameter $\gameParams$, the \emph{solution} of a \ac{gnep} is an equilibrium strategy profile $\mathbf{\traj}^* := (\traj^{1*}, \ldots, \traj^{N*})$ so that each agent's strategy is a best response to the others', \ie,
\begin{align}
   \traj^{i*} \in \mathcal{S}_\gameParams^i(\traj^{\neg i*}), \forall i \in [N].
\end{align}
Intuitively, at a \ac{gne}, no player can further reduce their cost by unilaterally adopting another feasible strategy.

\smallskip\noindent\textbf{Example: Online Game-Theoretic Motion Planning.}
This work focuses on applying games to online motion planning in interaction with other agents, such as the intersection scenario shown in \cref{fig:motivation}.
In this context, the strategy of agent~$i$,  $\traj_i$, represents a \emph{trajectory}---\ie, a sequence of states and inputs extended over a finite horizon---which is recomputed in a receding-horizon fashion.
This paradigm results in the game-theoretic equivalent of \ac{mpc}: \emph{\ac{mpgp}}.
The construction a trajectory \emph{game} largely follows the procedure used in single-agent trajectory \emph{optimization}: $g_\gameParams^i(\cdot)$ encodes input and state constraints including those enforcing dynamic feasibility and collision avoidance; %
and $J_\gameParams^i(\cdot)$ encodes the $i^\mathrm{th}$ agent's objective such as reference tracking.
Adopting a convention in which index~1 refers to the ego agent, the equilibrium solution of the game then serves two purposes simultaneously: the opponents' solution, $\traj^{\neg1*}$, serves as a game-theoretic \emph{prediction} of their behavior while the ego agent's solution, $\traj^{1*}$, provides the corresponding best response.

\smallskip\noindent\textbf{The Role of Game Parameters $\boldsymbol{\gameParams}$.}
A \emph{key difference} between trajectory games and single-agent trajectory optimization is the requirement to provide the costs and constraints of \emph{all} agents. %
In practice, a robot may have insufficient knowledge of their opponents' intents, dynamics, or states to instantiate a complete game-theoretic model.
This aspect motivates the parameterized formulation of the game above:
for the remainder of this manuscript, $\gameParams$ will capture the unknown aspects of the game.
In the context of game-theoretic motion planning, $\gameParams$ typically includes aspects of opponents' preferences, such as their unknown desired lane or preferred velocity. %
For conciseness, we denote game~(\ref{eq:forward-gnep}) compactly via the parametric tuple of problem data $\game(\gameParams) := (\{J^i_\gameParams, g^i_\gameParams\}_{i \in [N]})$.
Next, we discuss how to infer these parameters online from observed interactions.

\section{Formalizing Bayesian Inverse Games}\label{sec:problem-statement}
This section presents our main contribution: a framework for inferring unknown game parameters~$\gameParams$ based on observations~$\observation$.
This problem is referred to as an \emph{inverse} game~\cite{waugh2013computational}.

Several prior works on inverse games \cite{peters2023ijrr,li2023cost,clarke2023learning} seek to find game parameters that directly maximize observation likelihood.
However, this \ac{mle} formulation of inverse games (\romannumeral 1) only provides a point estimate of the unknown game parameters $\gameParams$, thereby precluding the consideration of uncertainty in downstream tasks such as motion planning; and (\romannumeral 2) fails to provide reasonable parameter estimates when observations are uninformative, as we shall also demonstrate in \cref{sec:exp}.

\subsection{A Bayesian View on Inverse Games}
\label{sec:bayesian-inverse-games}
In order to address the limitations of the \ac{mle} inverse games, we consider a \emph{Bayesian} formulation of inverse games, and seek to construct the belief distribution
\begin{align}\label{eq:bayesian-inverse-game}
b(\gameParams) = p(\gameParams \given \observation) = \frac{p(\observation \given \gameParams) p(\gameParams)}{p(\observation)}.
\end{align}
In contrast to \ac{mle} inverse games, this formulation provides a full posterior distribution over the unknown game parameters $\gameParams$ and factors in prior knowledge $p(\gameParams)$.

\smallskip\noindent\textbf{Observation Model $\boldsymbol{p(\observation\given\theta)}$.}
In autonomous online operation of a robot, $\observation$ represents the (partial) observations of other players' recent trajectories---\eg, from a fixed lag buffer as shown in orange in \cref{fig:motivation}.
Like prior works on inverse games~\cite{le2021lucidgames,li2023cost,peters2023ijrr}, we assume that, given the unobserved true trajectory of the human, $\observation$ is Gaussian-distributed; \ie, $p(\observation\given\traj) = \normaldist(\observation\given\mu_\observation(\traj),\Sigma_\observation(\traj))$.
Assuming that the underlying trajectory is the solution of a game with known structure $\game$ but unknown parameters $\gameParams$, we express the observation model as $p(\observation\given\gameParams) = p(\observation\given\solveGame(\gameParams))$, where $\solveGame$ denotes a game solver that returns a solution $\traj^*$ of the game $\game(\gameParams)$.

\smallskip\noindent\textbf{Challenges of Bayesian Inverse Games.}
While the Bayesian formulation of inverse games in \cref{eq:bayesian-inverse-game} is conceptually straightforward, it poses several challenges:
{
(\romannumeral1) The prior $p(\gameParams)$ is typically unavailable and instead must be learned from data.
(\romannumeral2) The computation of the normalizing constant, $p(\observation) = \int p(\observation \given \gameParams) p(\gameParams) \mathrm{d}\gameParams$, is intractable in practice due to the marginalization of $\gameParams$.
(\romannumeral3) Both the prior, $p(\gameParams)$, and posterior $p(\gameParams \given \observation)$ are in general non-Gaussian or even \emph{multi-modal} and are therefore difficult to represent explicitly in terms of their \ac{pdf}.
}
Prior work~\cite{le2021lucidgames} partially mitigates these challenges by using an \ac{ukf} for approximate Bayesian inference, but that approach is limited to unimodal uncertainty models and requires solving multiple games for a single belief update, thereby posing a computational challenge.

Fortunately, as we shall demonstrate in \cref{sec:exp}, many practical applications of inverse games do not require an explicit evaluation of the belief \ac{pdf}, $b(\gameParams)$.
Instead, a \emph{generative model} of the belief---\ie, one that allows drawing samples $\gameParams \sim b(\gameParams)$---often suffices.
Throughout this section, we demonstrate how to learn such a generative model from an unlabeled dataset $\dataset = \{\observation_k \mid \observation_k \sim p(\observation), \forall k \in [K]\}$ of observed interactions.

\subsection{Augmentation to Yield a Generative Model}\label{sec:augmentation}

\begin{figure}
    \centering
    \includegraphics[width=0.75\linewidth]{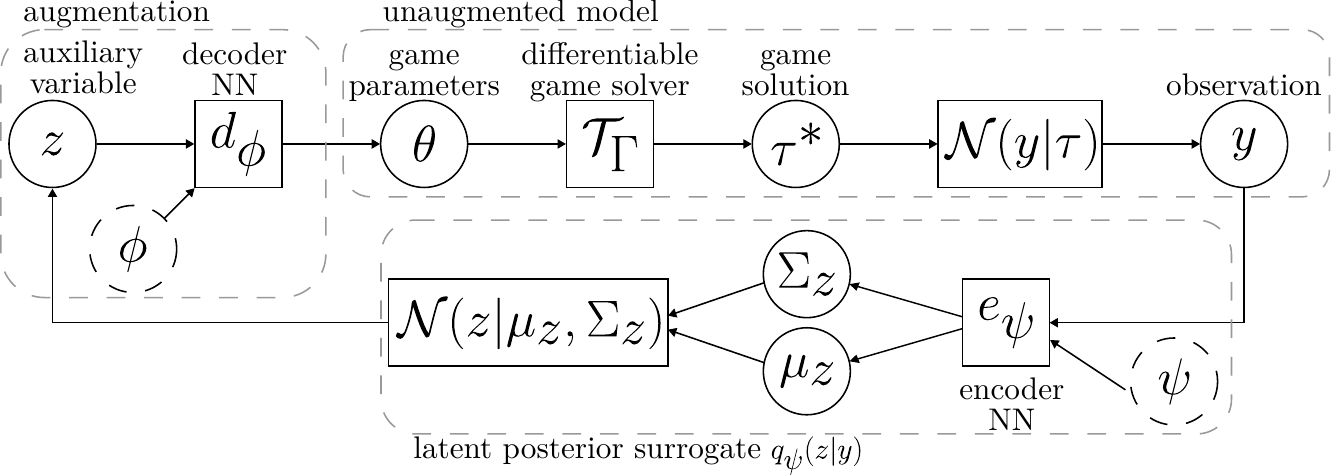}
    \caption{Overview of a structured \ac{vae} for generative Bayesian inverse games. Top (left to right): decoder pipeline. Bottom (right to left): variational inference via an encoder.}
    \label{fig:pipeline}
\end{figure}

To obtain a generative model of the belief $b(\gameParams)$, we augment our Bayesian model as summarized in the top half of \cref{fig:pipeline}.
The goal of this augmentation is to obtain a model structure that lends itself to approximate inference in the framework of \acfp{vae}~\cite{kingma2013auto}.
\revised{
As in a conventional \ac{vae}, we introduce an auxiliary random variable~$\latent$ with known isotropic Gaussian prior $p(\latent) = \normaldist(\latent\mid \mu=0,\Sigma=I)$ to model the data distribution as the marginal $p_\decoderParams(\observation) = \int p_\decoderParams(\observation\given\latent)p(\latent)\mathrm{d}\latent$.
We thus convert the task of learning the data distribution into learning the parameters of the conditional distribution $p_\decoderParams(\observation\given\latent)$, commonly referred to as the \emph{decoder} in terminology of \acp{vae}.
The key difference of our approach to a conventional \ac{vae} is the special \emph{structure} of this decoder.
Specifically, while a conventional \ac{vae} employs an (unstructured) \ac{nn}, say $\decoder$, to map $\latent$ to the parameters of the observation distribution, \ie $p_\decoderParams(\observation\given\latent) = p(\observation \given \decoder(\latent))$, our decoder composes this \ac{nn} with a game solver $\solveGame$ to model the data distribution as~$p(\observation\given (\solveGame\circ\decoder)(\latent))$.
From the perspective of a conventional \ac{vae}, the game solver thus takes the role of a special \say{layer} in the decoder, mapping game parameters predicted by $\decoder$ to the observation distribution.
In more rigorous probabilistic terminology, our proposed generative model can be interpreted as factoring the conditional distribution as $p_\decoderParams(\observation\given\latent) = \int p(\observation\given\gameParams)p_\decoderParams(\gameParams\given\latent)\mathrm{d}\gameParams$, with $p_\decoderParams(\gameParams\given\latent) = \delta(\gameParams - \decoder(\latent))$.
Here, $\delta$ denotes the Dirac delta function\footnote{\revised{
The modeling of $p_\decoderParams(\gameParams\given\latent)$ as a Dirac delta function---\ie a deterministic relationship between~$\latent$ and~$\gameParams$---is consistent with the common modeling choice in non-hierarchical \acp{vae}~\cite{kingma2013auto} which employ a \emph{deterministic} decoder network. 
We found good performance with this setup in our experiments; \cf, \cref{sec:exp}. We discuss extensions to other modeling choices in \cref{sec:conclusion}.
}}
and the observation model $p(\observation\given\gameParams)$ includes the game solver $\solveGame$ as discussed in~\cref{sec:bayesian-inverse-games}.
}
\vspace{-2em}
\subsection{Auto-Encoding Bayesian Inverse Games}\label{sec:structured-vae}
\revisedInline{
The modeling choices made in \cref{sec:augmentation} imply
that the game parameters~$\gameParams$ and observations~$\observation$ are conditionally independent given the latent variable~$\latent$.
Therefore, knowledge of the distribution of $\latent$ fully specifies the downstream distribution of $\gameParams$.
Specifically, observe that we can express the prior over the game parameters as $p_\decoderParams(\gameParams) = \int p_\decoderParams(\gameParams\given\latent)p(\latent)\mathrm{d}\latent$ and the corresponding posterior as $p_\decoderParams(\gameParams\given\observation) = \int p_\decoderParams(\gameParams\given\latent)p_\decoderParams(\latent\given\observation)\mathrm{d}\latent$.
}
\revisedInline{Intuitively, this means that} by propagating samples from the latent prior $p(\latent)$ through $\decoder$, we implicitly generate samples from the prior $p_\decoderParams(\gameParams)$. Similarly, by propagating samples from the latent posterior $p_\decoderParams(\latent\given\observation)$ through $\decoder$, we implicitly generate samples from the posterior $p_\decoderParams(\gameParams\given\observation)$.
We thus have converted the problem of (generative) Bayesian inverse games to estimation of $\decoderParams$ and inference of $p_\decoderParams(\latent \given \observation)$.

\revised{
As in conventional \acp{vae}, we can perform both these tasks through the lens of (amortized) \ac{vi}.
That is,
}
we seek to approximate the latent posterior $p_\decoderParams(\latent \given \observation)$ with a Gaussian $q_\encoderParams(\latent\given\observation) = \normaldist(\latent\given\encoder(\observation))$.
Here, $\encoder$ takes the role of an \emph{encoder} that maps observation~$\observation$ to the mean $\mu_{\latent}(\observation)$ and covariance matrix $\Sigma_{\latent}(\observation)$ of the Gaussian posterior approximation; \cf bottom of \cref{fig:pipeline}.
\revised{The design of this encoder model directly follows that of a conventional \ac{vae}~\cite{kingma2013auto} and can be realized via a \ac{nn}.}

\revised{
As is common in amortized \ac{vi}, we seek to fit the parameters of both the encoder and the structured decoder via gradient-based optimization.
A key challenge in taking this approach, however, is the fact that this training procedure requires back-propagation of gradients through the entire decoder; including the game solver $\solveGame$.
}
It is this gradient information from the game-theoretic \say{layer} in the decoding pipeline that induces an interpretable structure on the output of the decoder-\ac{nn} $\decoder$, forcing it to predict the hidden game parameters $\gameParams$. %
\revised{We recover the gradient of the game solver using the implicit differentiation approach proposed in~\cite{liu2023learning}.} %

\smallskip\noindent\textbf{Remark.} To generate samples from the estimated posterior $q_{\encoderParams,\decoderParams}(\gameParams \given \observation) = \int p_\decoderParams(\gameParams \given \latent)q_\encoderParams(\latent\given\observation) \mathrm{d}\latent$, we do not need to evaluate the game solver $\solveGame$: posterior sampling involves only the evaluation of \acp{nn} $\decoder$ and $\encoder$; \cf $\observation\to\gameParams$ in \cref{fig:pipeline}.

Next, we discuss the training of the \acp{nn}~$\decoder$ and~$\encoder$ of our structured \ac{vae}.

\subsection{Training the Structured Variational Autoencoder}\label{sec:training}
Below, we outline the process for optimizing model parameters in our specific setting.
For a general discussion of \acp{vae} and \ac{vi}, we refer to \cite{murphy2023probabilistic}. %

\smallskip\noindent\textbf{Fitting a Prior to $\dataset$.}
First, we discuss the identification of the prior parameters $\decoderParams$.
We fit these parameters so that the data distribution induced by $\decoderParams$, \ie, $p_\decoderParams(\observation) = \int p_\decoderParams(\observation \given \latent) p(\latent) \mathrm{d}\latent$, closely matches the unknown true data distribution $p(\observation)$.
For this purpose, we measure closeness between distributions via the \ac{kl} divergence
\begin{align}
    \dkl{p}{q} &:= \expectedValue{x\sim p(x)}{\log p(x) - \log q(x)}.
\end{align}
The key properties of this divergence metric, $\dkl{p}{q} \geq 0$ and $\dkl{p}{q} = 0 \iff p = q$, allow us to cast the estimation of $\decoderParams$ as an optimization problem: %
\begin{subequations}
\label{eq:prior-dkl}
\begin{align}
    \decoderParams^*
    &&\in \arg\min_\decoderParams \quad& \dkl{p(\observation)}{p_\decoderParams(\observation)}\\
    &&= \arg\min_\decoderParams \quad& -\expectedValue{\observation \sim p(\observation)}{\log \expectedValue{\latent\sim p(\latent)}{p_\decoderParams(\observation \given \latent)}}.\label{eq:prior-dkl-simplified}
\end{align}
\end{subequations}
With the prior recovered, we turn to the approximation of the posterior $p_\decoderParams(\gameParams\given\observation)$.

\smallskip\noindent\textbf{Variational Belief Inference.}
We can find the closest surrogate $q_\encoderParams(\latent\given\observation)$ of $p_\decoderParams(\latent\given\observation)$ by minimizing the expected \ac{kl} divergence between the two distributions over the data distribution $\observation \sim p(\observation)$ using the framework of \ac{vi}:
\begin{subequations}
\label{eq:belief_vi}
\begin{align}
    \encoderParams^* &\in \arg\min_\encoderParams \expectedValue{\observation\sim p(\observation)}{\dkl{q_\encoderParams(\latent\given\observation)}{p_\decoderParams(\latent\given\observation)}}\label{eq:belief_vi_dkl}\\
                     &= \arg\min_\encoderParams \expectedValue{\substack{\observation\sim p(\observation)\\\latent\sim q_\encoderParams(\latent\given\observation)
                 }}{\losssample(\encoderParams,\decoderParams,\observation,\latent)}\label{eq:belief_vi_final},
\end{align}
\end{subequations}
\begin{align*}\label{eqn:losssample}
    \text{where}\quad \losssample(\encoderParams,\decoderParams,\observation,\latent) := \log \underbrace{q_\encoderParams(\latent\given\observation)}_{\normaldist(\latent\given\encoder(\observation))} - \log \underbrace{p_\decoderParams(\observation\given\latent)}_{\normaldist(\observation\given(\solveGame\circ\decoder)(\latent))} - \log \underbrace{p(\latent)}_{\normaldist(\latent)}.
\end{align*}
We have therefore outlined, in theory, the methodology to determine all parameters of the pipeline from the unlabeled dataset $\dataset$.

\smallskip\noindent\textbf{Considerations for Practical Realization.}
To solve \cref{eq:prior-dkl,eq:belief_vi} in practice, we must overcome two main challenges:
first, the loss landscapes are highly nonlinear %
due to the nonlinear transformations $\encoder,\decoder$ and $\solveGame$;
and second, the expected values cannot be computed in closed form.
These challenges can be addressed by taking a \emph{stochastic} first-order optimization approach, such as \ac{sgd}, %
thereby limiting the search to a \emph{local} optimum and approximating the gradients of the expected values via Monte Carlo sampling.
To facilitate \ac{sgd}, we seek to construct unbiased gradient estimators of the objectives of \cref{eq:prior-dkl-simplified} and \cref{eq:belief_vi_final} from gradients computed at individual samples.
An unbiased gradient estimator is straightforward to define when the following conditions hold: (\gradcondition{1}) any expectations appear on the outside, and (\gradcondition{2}) the sampling distribution of the expectations are independent of the variable of differentiation.
The objective of the optimization problem in \cref{eq:belief_vi_final} takes the form
\begin{align}
    \loss(q, \decoderParams)
    := \expectedValue{\substack{\observation\sim p(\observation)\\\latent\sim q(\latent\given\observation)}}{{
    \losssample(\encoderParams,\decoderParams,\observation,\latent)
   }},
\end{align}
which clearly satisfies \gradcondition{1}.
Similarly, it is easy to verify that the objective of \cref{eq:prior-dkl-simplified} can be identified as $\loss(p_\decoderParams(\latent\given\observation), \cdot)$. %
Unfortunately, this latter insight is not immediately actionable since $p_\decoderParams(\latent\given\observation)$ is not readily available.
However, if instead we use our surrogate model from \cref{eq:belief_vi}---which, by construction closely matches $p_\decoderParams(\latent\given\observation)$---we can cast the estimation of the prior and posterior model as a joint optimization of $\loss$:
\begin{align}\label{eq:joint-vi}
\tilde{\encoderParams}^*,\tilde{\decoderParams}^* \in \arg\min_{\encoderParams,\decoderParams} \loss(q_\encoderParams,\decoderParams).
\end{align}
When $\encoder$ and $\decoder$ are sufficiently expressive to allow $\dkl{q_\encoderParams(\latent\given\observation)}{p_\decoderParams(\latent\given\observation)} = 0$, then this reformulation is exact; \ie, $\tilde{\decoderParams}^*$ and $\tilde{\encoderParams}^*$ are also minimizers of the original problems \cref{eq:prior-dkl} and \cref{eq:belief_vi}, respectively.
In practice, a perfect match of distributions is not typically achieved and $\tilde{\decoderParams}^*$ and $\tilde{\encoderParams}^*$ are biased.
Nonetheless, the scalability enabled by this reformulation has been demonstrated to enable generative modeling of complex distributions, including those of real-world images~\cite{gulrajani2016pixelvae}. %
Finally, to also satisfy \gradcondition{2} for the objective of \cref{eq:joint-vi}, we apply the well-established \say{reparameterization trick} to cast the inner expectation over a sampling distribution independent of $\encoderParams$.
That is, we write
\begin{align}\label{eq:joint-vi-reparameterized}
    \loss(q_\encoderParams,\decoderParams)
    &= \expectedValue{\substack{\observation\sim p(\observation)\\\noise\sim\normaldist}}{
        \losssample(\encoderParams,\decoderParams,\observation,\remap_{q_\encoderParams}(\noise, \observation))
    },
\end{align}
where the inner expectation is taken over $\noise$ that has multi-variate standard normal distribution, and $\remap_{q_\encoderParams}(\cdot,\observation)$ defines a bijection from $\noise$ to $\latent$ so that $\latent\sim q_\encoderParams(\cdot\given\observation)$.
Since we model the latent posterior as Gaussian---\ie, $q_\encoderParams(\latent\given\observation)=\normaldist(\latent\given\encoder(\observation))$---the reparameterization map is $\remap_{q_\encoderParams}(\noise, \observation):= \mu_{\latent}(\observation) + L_{\latent}(\observation)\noise$, where $L_\latent L_\latent^\transpose = \Sigma_\latent$ is a Cholesky decomposition of the latent covariance.
Observe that \cref{eq:joint-vi-reparameterized} only involves Gaussian \acp{pdf}, and all terms can be easily evaluated in closed form.

\smallskip\noindent\textbf{Stochastic Optimization.}
As in \ac{sgd}-based training of conventional \acp{vae}, at iteration~$k$, we sample~$\observation_k\sim\dataset$, and encode~$\observation_k$ into the parameters of the latent distribution via~$\encoder[\encoderParams_k]$.
From the latent distribution we then sample $\latent_k$, and evaluate the gradient estimators $\nabla_\encoderParams\losssample(\encoderParams,\decoderParams_k,\observation_k,\latent_k)|_{\encoderParams = \encoderParams_k}$ and $\nabla_\decoderParams\losssample(\encoderParams_k,\decoderParams,\observation_k,\latent_k)|_{\decoderParams=\decoderParams_k}$ of $\loss$ at the current parameter iterates, $\encoderParams_k$ and $\decoderParams_k$. %
Due to our model's structure, the evaluation of these gradient estimators thereby also involves the differentiation of the game solver $\solveGame$. %

\section{Experiments}
\label{sec:exp}

To assess the proposed approach, we evaluate its online inference capabilities and its efficacy in downstream motion planning tasks in several simulated driving scenarios. 
Experiment videos can be found in our \href{https://xinjie-liu.github.io/projects/bayesian-inverse-games}{supplementary video: \color{magenta}https://xinjie-liu.github.io/projects/bayesian-inverse-games}.

\vspace{-0.5em}
\subsection{Planning with Online Inference}
\label{two-player-intersection}
First, we evaluate the proposed framework for downstream motion planning tasks in the two-player intersection scenario depicted in~\cref{fig:intersection-qualitative}.
This experiment is designed to validate the following hypotheses:
\begin{itemize}
    \item \textbf{\textbf{H1} (Inference Accuracy).}  Our method provides more accurate inference than \ac{mle} by leveraging the learned prior information, especially in settings where multiple human objectives explain the observations.
    \item \textbf{\textbf{H2} (Multi-modality).}  Our approach predicts posterior distributions that capture the multi-modality of agents' objectives and behavior. 
    \item \textbf{\textbf{\textbf{H3}} (Planning Safety).} Bayesian inverse game solutions enable safer robot plans compared to \ac{mle} methods.
\end{itemize}
\vspace{-1.5em}
\subsubsection{Experiment Setup}
In the test scenario shown in~\cref{fig:intersection-qualitative}, the red robot navigates an intersection while interacting with the green human, whose goal position is initially unknown.
In each simulated interaction, the human's goal is sampled from a Gaussian mixture with equal probability for two mixture components: one for turning left, and one for going straight.
\revised{To simulate human behavior that responds strategically to the robot's decisions, we generate the human's actions by solving trajectory games in a receding-horizon fashion with access to the sampled ground-truth game parameters.}

We model the agents' dynamics as kinematic bicycles.
The state at time step~$t$ includes position,
longitudinal velocity, and orientation, \ie, $x_t^i = (p_{x,t}^i, p_{y,t}^i, v_{t}^i, \xi_{t}^i)$, and the control comprises acceleration and steering angle, \ie, $u_t^i = (a_t^i, \eta_t^i)$.
We assign player index~$1$ to the robot and index~$2$ to the human.
Over a planning horizon of ~$T = 15$ time steps, each player~$i$ seeks to minimize a cost function that encodes incentives for reaching the goal, reducing control effort, and avoiding collision:
\begin{multline}
    \label{eq: intersection cost structure}
    \cost^i_{\costparams} = \sum_{t = 1}^{T-1} \|p_{t+1}^i-p^i_{\textrm{goal}}\|_2^2 +  0.1 \| u_t^i \|_2^2
     + 400 \max(0, d_{\min} - \|p_{t+1}^i - p_{t+1}^{\neg i}\|_2)^3,
\end{multline}
\noindent where~$p^i_t = (p^i_{x,t}, p^i_{y,t})$ denotes agent~$i$'s position at time step $t$, $p^i_{\textrm{goal}} = (p^i_{\textrm{goal}, x}, p^i_{\textrm{goal}, y})$ denotes their goal position, and $d_\mathrm{min}$ denotes a preferred minimum distance to other agents.
The unknown parameter~$\theta$ inferred by the robot contains the two-dimensional goal position~$p_{\textrm{goal}}^2$ of the human.

We train the structured~\ac{vae} from~\cref{sec:structured-vae} on a dataset of observations from~560 closed-loop interaction episodes obtained by solving dynamic games with the opponent's ground truth goals sampled from the Gaussian mixture described above.
The 560 closed-loop interactions are sliced into~34600~15-step observations $\observation$.
Partial-state observations consist of the agents' positions and orientations. 
We employ fully-connected feedforward \acp{nn} with two 128-dimensional and 80-dimensional hidden layers as the encoder model~$\encoder$ and decoder model~$\decoder$, respectively.
The latent variable~$\latent$ is~16-dimensional. 
We train the \ac{vae} with Adam~\cite{kingma2014adam} for 100 epochs, which took 14 hours of wall-clock time.

\subsubsection{Baselines}
We evaluate the following methods to control the robot interacting with a human of unknown intent\revised{, where both the robot and the human trajectories are computed game-theoretically at every time step in a receding-horizon fashion}: %

\emph{Ground truth (GT)}: This planner generates robot plans by solving games with access to ground truth opponent objectives. %
We include this oracle baseline to provide a reference upper-bound on planner performance.

\emph{Bayesian inverse game (ours) + planning in expectation (B-PinE)}
solves multi-hypothesis games~\cite{peters2024ral} in which the robot minimizes the \emph{expected} cost~$\expectedValue{\costparams \sim  q_{\encoderParams,\decoderParams}(\gameParams \given \observation)}{J_{\costparams}^1}$ under the posterior distributions predicted by our structured \ac{vae} based on new observations at each time step.
    
\emph{ Bayesian inverse game (ours) + \ac{map} planning (B-MAP)} constructs \ac{map} estimates~$\hat{\costparams}_{\mathrm{MAP}} \in \arg\max_\gameParams  q_{\encoderParams,\decoderParams}(\gameParams \given \observation)$ from the posteriors and solves the game $\game(\hat{\gameParams}_\mathrm{MAP})$.
    
\emph{Randomly initialized \ac{mle} planning (R-MLE)}: This baseline~\cite{liu2023learning} solves the game $\game(\hat{\gameParams}_\mathrm{MLE})$, where~$\hat{\costparams}_{\mathrm{MLE}} \in \arg\max_\gameParams p(\observation \given \gameParams)$.
\revised{Hence, this baseline utilizes the \emph{same game structure} as our method but only makes a point estimate of the game parameters.}
The \ac{mle} problems are solved online via gradient descent based on new observations at each time step as in~\cite{liu2023learning}.
The initial guess for optimization is sampled uniformly from a rectangular region covering all potential ground truth goal positions. 

\emph{ Bayesian prior initialized \ac{mle} planning (BP-MLE)}: Instead of uniformly sampling heuristic initial guesses as in R-MLE, this baseline solves for the \ac{mle} with initial guesses from the Bayesian \textit{prior} learned by our approach.
    
\emph{Static Bayesian prior planning (St-BP)} samples~$\hat{\gameParams}$ from the learned Bayesian \emph{prior} and uses the sample as a \textit{fixed} human objective estimate to solve~$\game(\hat{\gameParams})$.
    This baseline is designed as an ablation study of the effect of online objective inference.

For those planners that utilize our \ac{vae} for inference, we take 1000 samples at each time step to approximate the distribution of human objectives, which takes around \SI{7}{ms}.
For B-PinE, we cluster the posterior samples into two groups to be compatible with the multi-hypothesis game solver from~\cite{peters2024ral}.
\vspace{-0.5em}
\subsubsection{Qualitative Behavior}
\label{sec:intersection-qualitative}
\begin{figure*}[h]
  \centering
  \includegraphics[width=\linewidth]{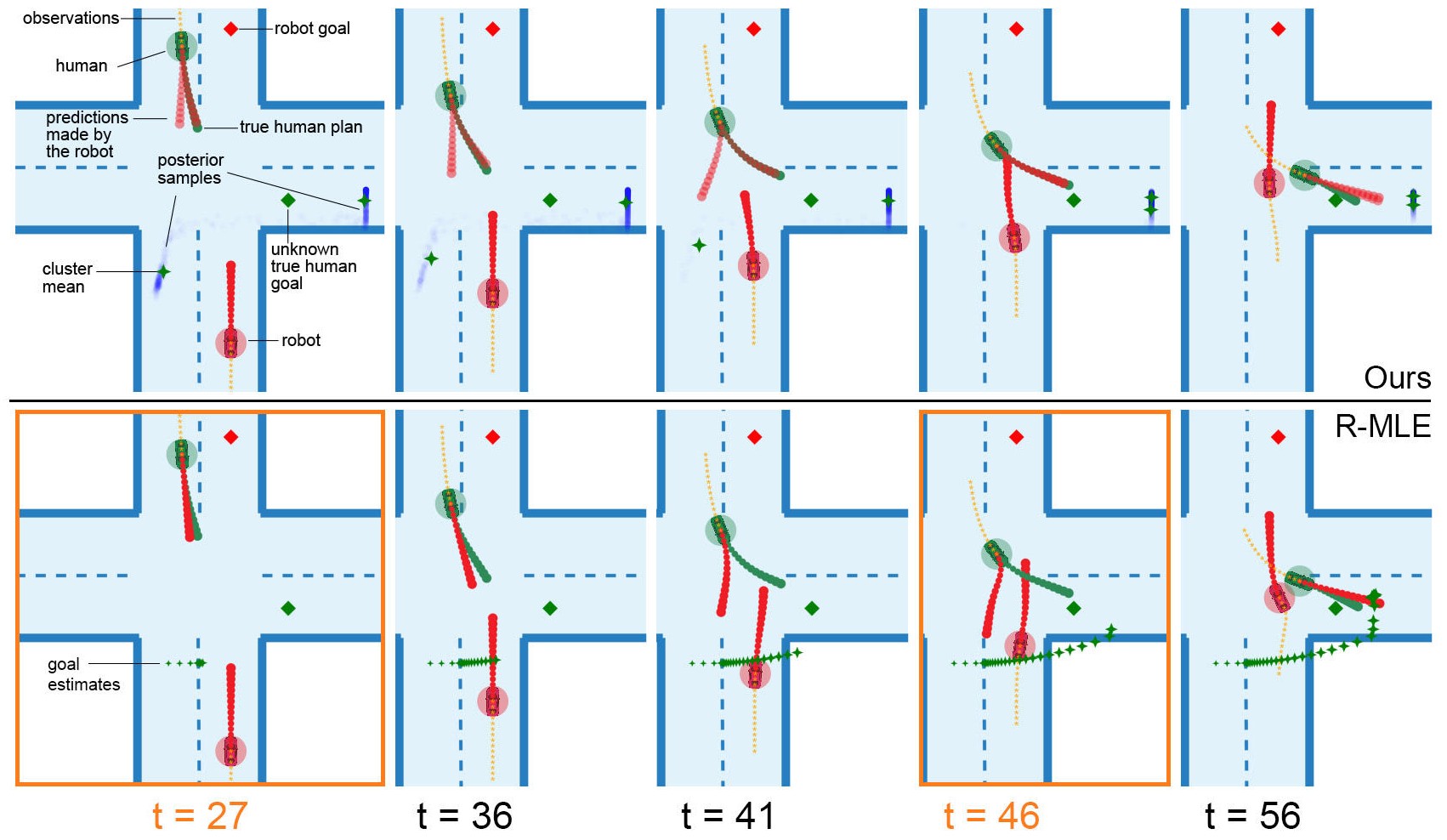}
  \caption{Qualitative behavior of B-PinE (ours) vs. R-MLE. 
  In the bottom row, the size of the green stars increases with time.
  }
  \label{fig:intersection-qualitative}
\end{figure*}

\Cref{fig:intersection-qualitative} shows the qualitative behavior of B-PinE and R-MLE.

\emph{B-PinE:} The top row of~\cref{fig:intersection-qualitative} shows that our approach initially generates a bimodal belief, capturing the %
distribution of potential opponent goals.
In the face of this uncertainty, the planner initially computes a more conservative trajectory to remain safe.
As the human approaches the intersection and reveals its intent, %
the left-turning mode gains probability mass until the computed belief eventually collapses to a unimodal distribution. %
Throughout the interaction, the predictions generated by the multi-hypothesis game accurately cover the true human plan, allowing safe and efficient interaction.

\emph{R-MLE:} 
As shown in the bottom row of~\cref{fig:intersection-qualitative}, the R-MLE baseline initially estimates that the human will go straight.
The true human goal only becomes clear later in the interaction and the over-confident plans derived from these poor point estimates eventually lead to a collision.

\begin{figure}[h]
  \centering
  \includegraphics[width=0.6\linewidth]{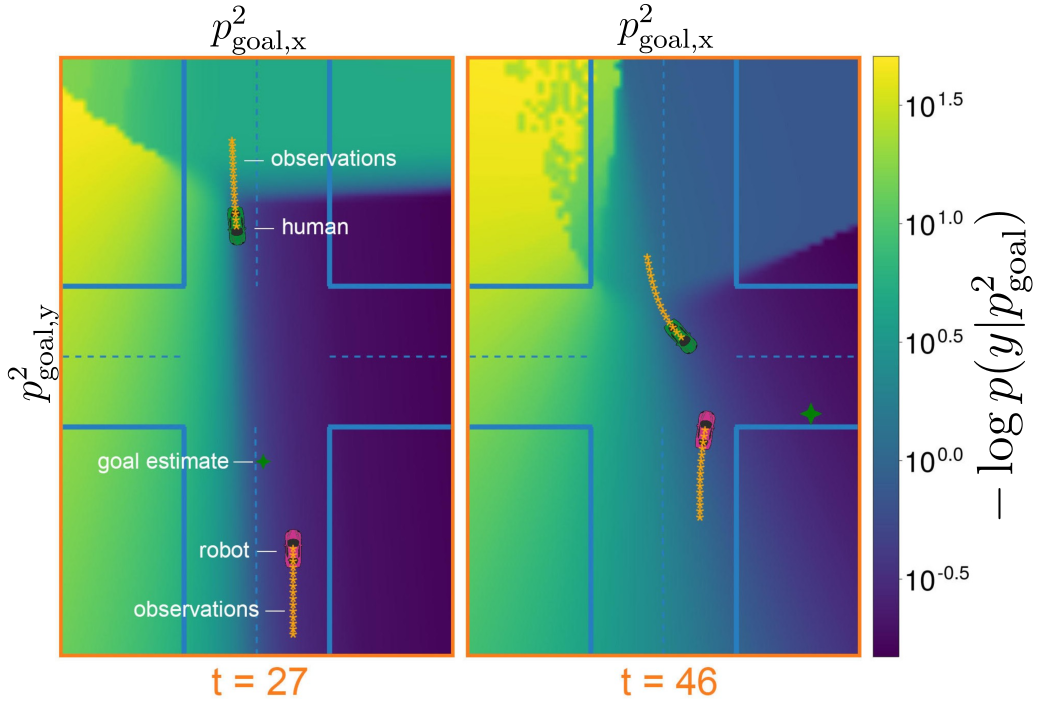}
  \caption{Negative observation log-likelihood~$-\log p(\observation \given p^2_{\mathrm{goal}})$ for varying human goal positions~$p^2_{\mathrm{goal}}$ at two time steps of the R-MLE trial in~\cref{fig:intersection-qualitative}. 
  }
  \label{fig:mle-cost}
\end{figure}
\emph{Observability Issues of MLE:}
To illustrate the underlying issues that caused the poor performance of the R-MLE baseline discussed above, \cref{fig:mle-cost} shows the negative observation log-likelihood for two time steps of the interaction.
As can be seen, a large region in the game parameter space explains the observed behavior well, and the baseline incorrectly concludes that the opponent's goal is always in front of their current position; \cf bottom of \cref{fig:intersection-qualitative}.
In contrast, by leveraging the learned prior distribution, our approach predicts objective samples that capture 
potential opponent goals well even when the observation is uninformative; \cf top of \cref{fig:intersection-qualitative}.

\textit{In summary, these results validate hypotheses \textbf{\textbf{H1}} and \textbf{\textbf{H2}}.}

\subsubsection{Quantitative Analysis}
To quantify the performance of all six planners, we run a Monte Carlo study of 1500 trials.
In each trial, we randomly sample the robot's
initial position along the lane from a uniform distribution such that the resulting ground truth interaction covers a spectrum of behaviors, ranging from the robot entering the intersection first to the human entering the intersection first.
We use the minimum distance between players in each trial to measure safety and the robot's cost as a metric for efficiency.

We group the trials into three settings based on the ground-truth behavior of the agents: \textbf{\textbf{(S1)}} The human turns left and passes the intersection first.
In this setting, a robot recognizing the human's intent should yield, whereas
blindly optimizing the goal-reaching objective will likely lead to unsafe interaction. 
\textbf{\textbf{(S2)}} The human turns left, but the robot reaches the intersection first.
In this setting, we expect the effect of inaccurate goal estimates to be less pronounced than in \textbf{S1} since the robot passes the human on the right irrespective of their true intent.
\textbf{\textbf{(S3)}} The human drives straight. Safety is trivially achieved in this setting.
\begin{figure*}[h]
  \centering
  \includegraphics[width=0.8\linewidth]{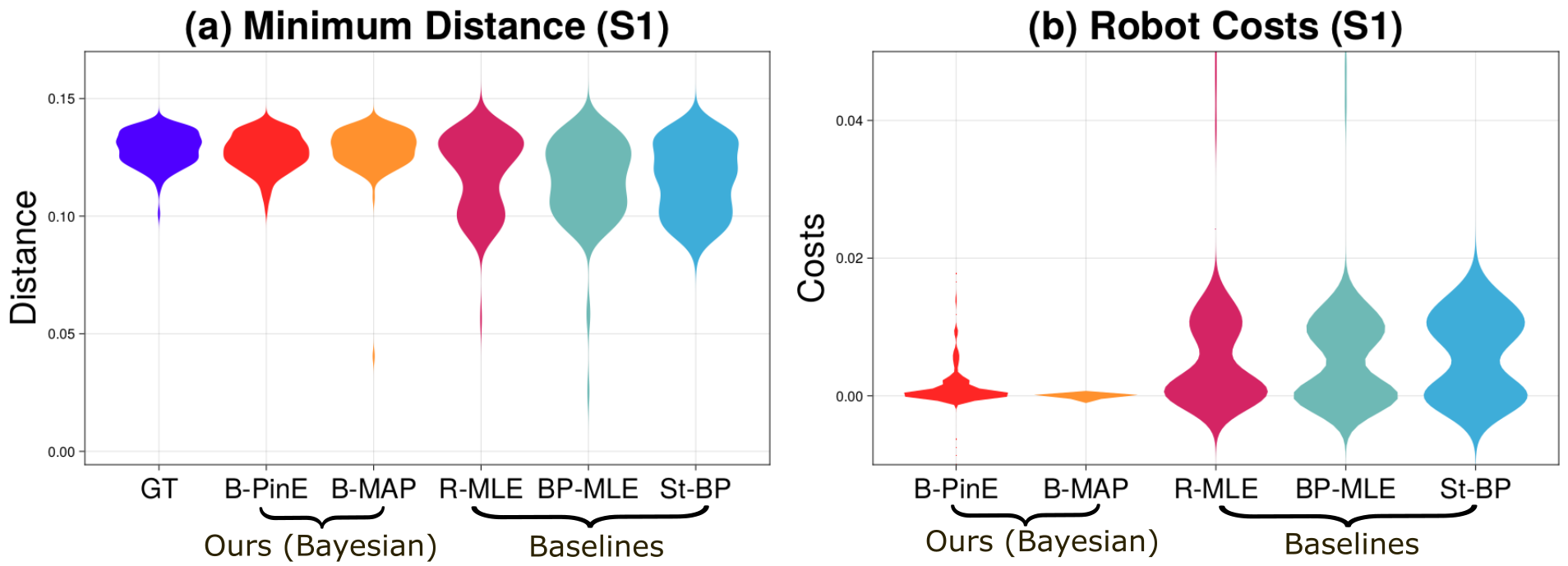}
  \caption{Quantitative results of \textbf{S1}: (a) Minimum distance between agents in each trial. (b) Robot costs (with GT costs subtracted).}
  \label{fig:intersection-quantitative-ego-second}
\end{figure*}
\begin{figure*}[h]
  \centering
  \includegraphics[width=\linewidth]{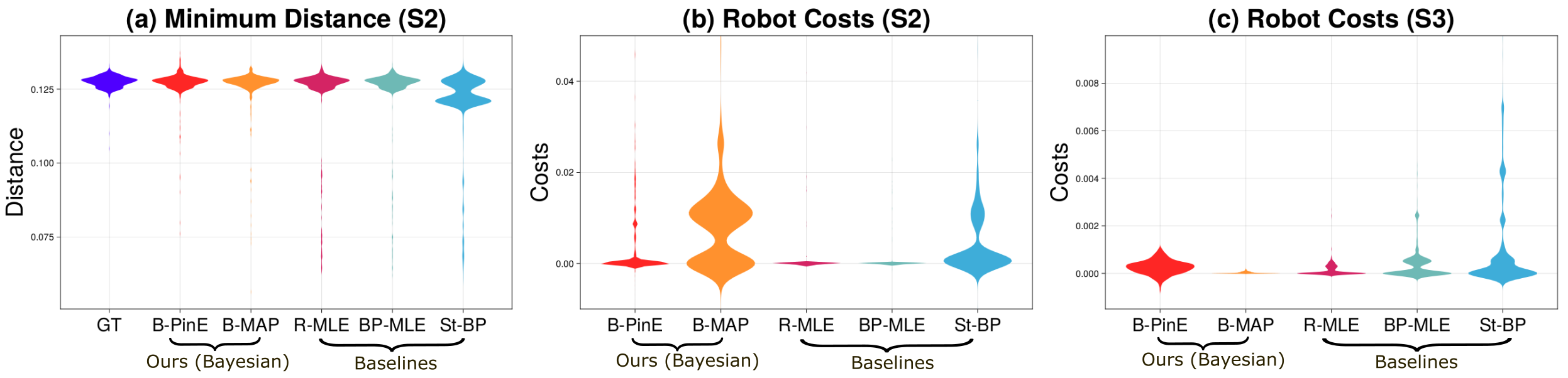}
  \caption{Quantitative results of \textbf{S2} and \textbf{S3}: (a) Minimum distance between agents in each trial of \textbf{S2}. (b-c) Robot costs of \textbf{S2-3} (with GT costs subtracted).
  }
  \label{fig:intersection-quantitative-ego-first}
\end{figure*}

\emph{Results, \textbf{S1}:}~\Cref{fig:intersection-quantitative-ego-second}(a) shows that methods using our framework achieve better planning safety than other baselines, closely matching the ground truth in this metric.
Using the minimum distance between agents among all ground truth trials as a collision threshold, the collision rates of the methods are \textbf{0.0\% (B-PinE)}, 0.78\% (B-MAP), 17.05\% (R-MLE), 16.28\% (BP-MLE), and 17.83\% (St-BP), respectively. 
Moreover, the improved safety of the planners using Bayesian inference does not come at the cost of reduced planning efficiency. 
As shown in ~\cref{fig:intersection-quantitative-ego-second}(\revised{b}), B-PinE and B-MAP consistently achieve low interaction costs.

\emph{Results, \textbf{S2}:} %
\Cref{fig:intersection-quantitative-ego-first}(a) shows that all the approaches except for the St-BP baseline approximately achieve ground truth safety in this less challenging setting.
While the gap between approaches is less pronounced, we still find improved planning safety of our methods over the baselines.
Taking the same collision distance threshold as in \textbf{S1}, the collision rates of the methods are \textbf{0.86\% (B-PinE)}, 2.24\% (B-MAP), 7.59\% (R-MLE), 6.03\% (BP-MLE), and 7.59\% (St-BP), respectively. 
B-MAP gives less efficient performance in this setting.
Compared with B-PinE, B-MAP only uses point estimates of the unknown goals and commits to over-confident and more aggressive behavior.
We observe that this aggressiveness results in coordination issues when the two agents enter the intersection nearly simultaneously, causing them to accelerate simultaneously and then brake together.
Conversely, the uncertainty-aware B-PinE variant of our method does not experience coordination failures, highlighting the advantage of incorporating the inferred \emph{distribution} during planning.
Lastly, St-BP exhibits the poorest performance in \textbf{S2}, stressing the importance of online objective inference.

\emph{Results, \textbf{S3}:} In~\cref{fig:intersection-quantitative-ego-first}(\revised{c}), B-MAP achieves the highest efficiency, while B-PinE produces several trials with increased costs due to more conservative planning.
Again, the St-BP baseline exhibits the poorest performance.

\emph{Summary:} These quantitative results show that our Bayesian inverse game approach improves motion planning safety over the \ac{mle} baselines, validating hypothesis \textbf{ \textbf{H3}}.
Between B-PinE and B-MAP, we observe improved safety in \textbf{S1-2} and efficiency in \textbf{S2}.

\subsection{Additional Evaluation of Online Inference}
\label{sec:inference-results}

To isolate inference effects from planning, we consider two additional online inference settings.
In these settings, agents operate with fixed but unknown ground-truth game parameters.
Our method acts as an external observer, initially aware only of the robots' intent, and is tasked to infer the humans' intent online.

\subsubsection{Two-Player Highway Game}

This experiment is designed to validate the following hypotheses:
\begin{itemize}
    \item \textbf{\textbf{H4} (Bayesian Prior).} Our method learns the underlying multi-modal prior objective distribution. 
    \item \textbf{\textbf{H5} (Uncertainty Awareness).} Our method computes a narrow, unimodal belief when the unknown objective is clearly observable 
    and computes a wide, multi-modal belief that is closer to the prior in case of uninformative observations. 
\end{itemize}
We consider a two-player highway driving scenario, in which an ego robot drives in front of a human opponent whose desired driving speed is unknown.
The rear vehicle is responsible for decelerating and avoiding collisions, and the front vehicle always drives at their desired speed. 
To simplify the visualization of beliefs and thereby illustrate \textbf{\textbf{H4}-5}, we reduce the setting to a single spatial dimension and model one-dimensional uncertainty: we limit agents to drive in a single lane and model the agents' dynamics as double integrators with longitudinal position and velocity as states and acceleration as controls.
Furthermore, we employ a \ac{vae} with the same hidden layers as in~\cref{two-player-intersection} but only \emph{one-dimensional latent space}.
The \ac{vae} takes a 15-step observation of the two players' velocities as input for inference.

\begin{figure}[h]
        \begin{subfigure}{0.5\textwidth}
            \includegraphics[width=0.8\linewidth]{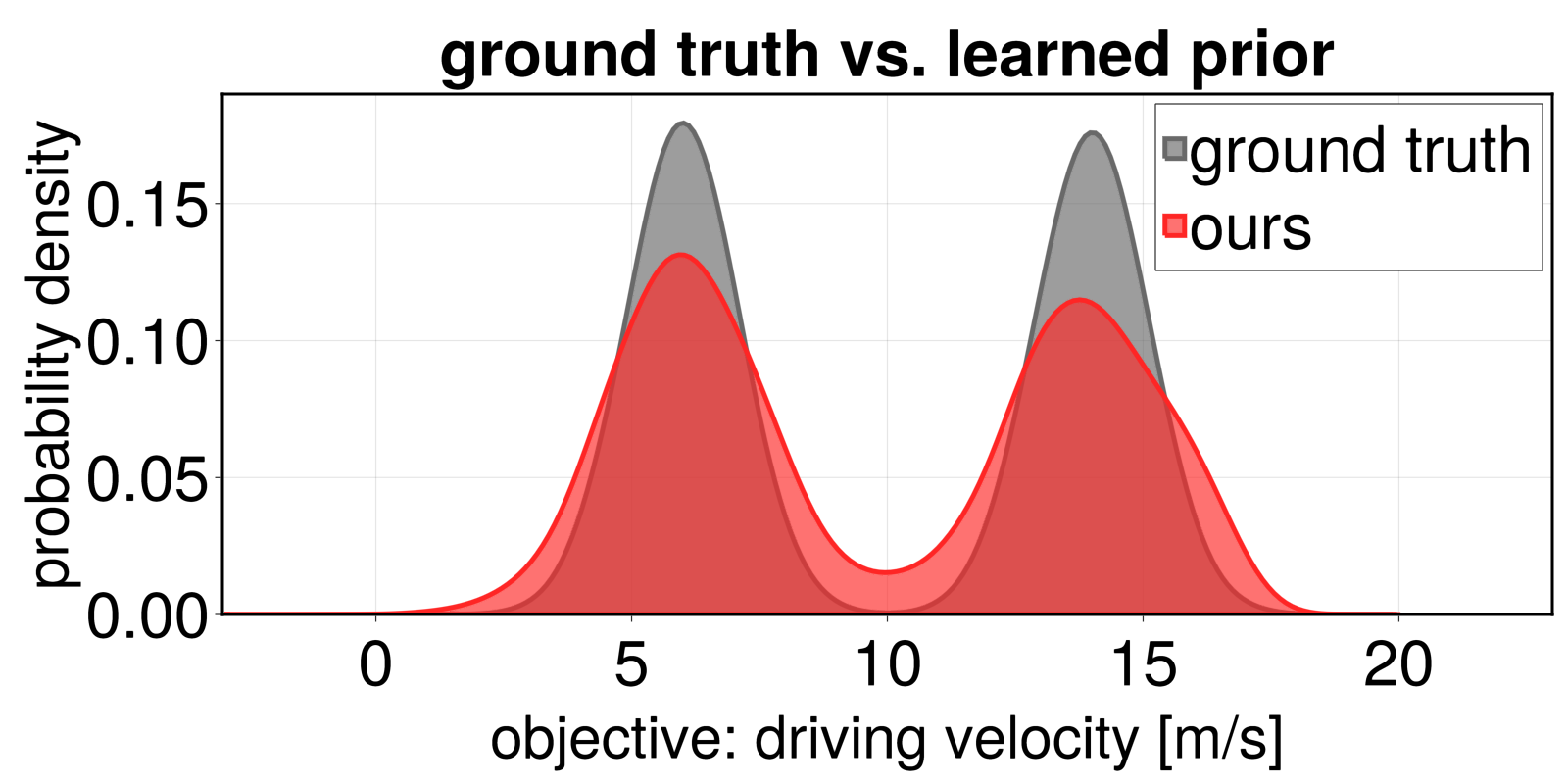}
            \caption{Prior distributions.}
        \end{subfigure}
        \hspace*{-0.8cm}
        \begin{subfigure}{0.55\textwidth}
		\includegraphics[width=1.0\linewidth]{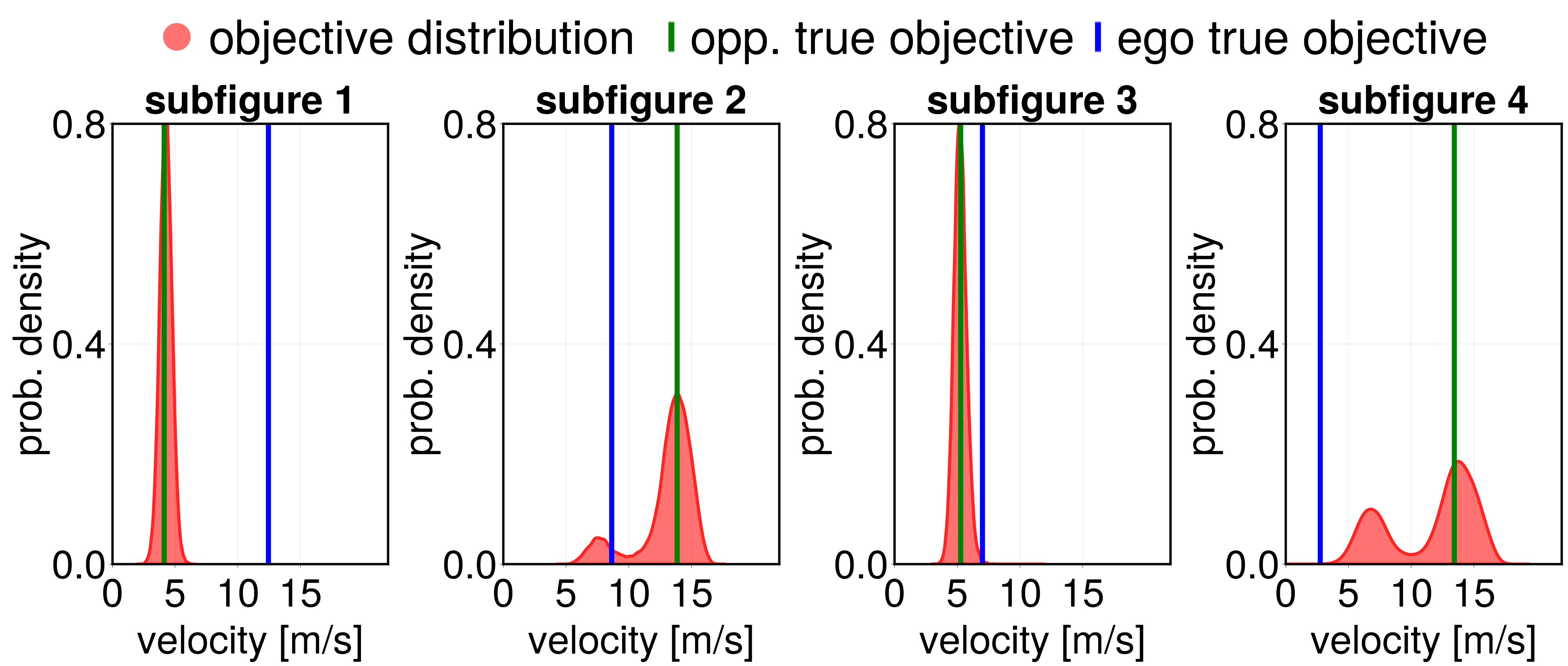}
            \caption{Posterior distributions.}
	\end{subfigure}

    \caption{(a) Learned and ground truth priors for the human's objective. (b) Inferred objective posterior distributions.}
    \label{fig:highway-results}
\end{figure}

We collect a dataset of 20000 observations to train a \ac{vae}. In each trial, the robot's reference velocity is sampled from a uniform distribution from $\SI{0}{\meter\per\second}$ to the maximum velocity of $\SI{20}{\meter\per\second}$; the human's desired velocity is sampled from a bi-modal Gaussian mixture distribution shown in grey in~\cref{fig:highway-results}(a) with two unit-variance mixture components at means of 30\% and 70\% of the maximum velocity. 

\Cref{fig:highway-results}(a) shows that the learned prior objective distribution captures the underlying training set distribution closely and thereby validates our hypothesis \textbf{\textbf{H4}}.
\Cref{fig:highway-results}(b) shows beliefs inferred by our approach for selected trials. 
The proposed Bayesian inverse game approach recovers a narrow distribution with low uncertainty when the opponent's desired speed is clearly observable,
\eg, when the front vehicle drives faster so that the rear vehicle can drive at their desired speed (\cref{fig:highway-results}, subfigures 1, 3).
The rear driver's objective becomes unobservable when they wish to drive fast but are blocked by the car in front.
In this case, our approach infers a wide, multi-modal distribution with high uncertainty that is closer to the prior distribution (\cref{fig:highway-results}, subfigures 2, 4). 
These results validate hypothesis \textbf{H5}.%

\subsubsection{Multi-Player Demonstration}
\begin{figure*}[h]
  \centering
  \includegraphics[width=0.325\linewidth]{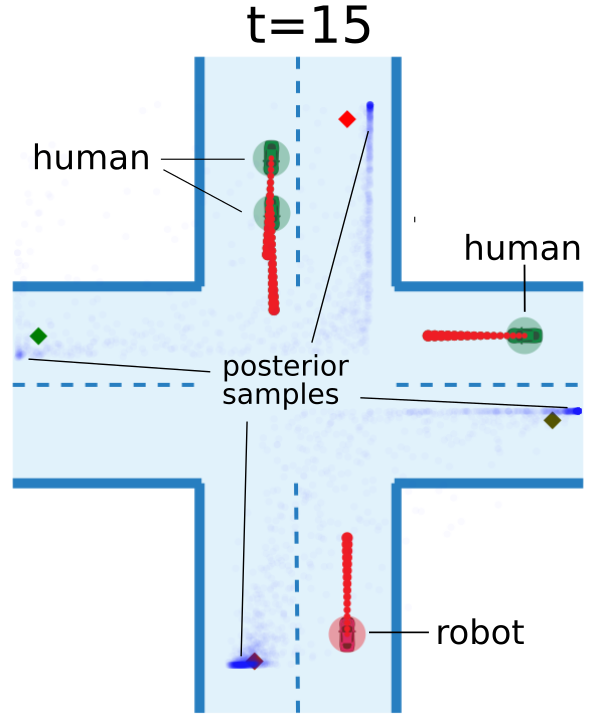}
  \includegraphics[width=0.32\linewidth]{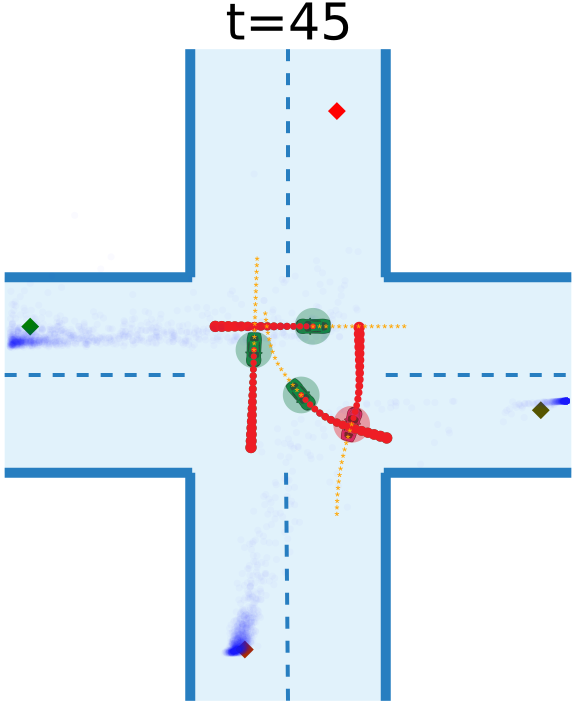}
  \includegraphics[width=0.32\linewidth]{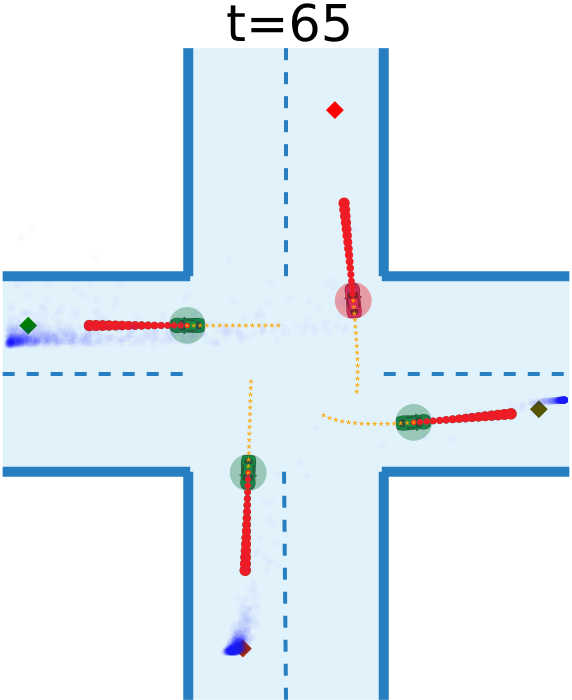}
  \caption{Belief evolution of our structured \ac{vae} in a 4-player intersection scenario.%
  }
  \label{fig:multi-agent-demo}
\end{figure*}

To demonstrate the scalability of our method, we tested our inference approach in a four-agent intersection scenario. As shown in~\cref{fig:multi-agent-demo}, a red robot interacts with three green humans, and our structured \ac{vae} infers the humans' goal positions. Initially, our method predicts multi-modal distributions with high uncertainty about the humans' objectives. As more evidence is gathered, uncertainty decreases, and the belief accurately reflects the true objectives of the other agents. The objective distribution inference operates in real-time at approximately \SI{150}{\hertz} without requiring additional game solves.
\vspace{-0.5em}

\section{Conclusion \& Future Work}
\label{sec:conclusion}
We presented a tractable Bayesian inference approach for dynamic games using a structured \ac{vae} with an embedded differentiable game solver.
Our method learns prior game parameter distributions from unlabeled data and infers multi-modal posteriors, outperforming MLE approaches in simulated driving scenarios.

Future work could explore this pipeline with other planning frameworks, incorporating active information gathering and intent signaling.
Furthermore, in this work we employed a decoder architecture that assumes a deterministic relationship between the latent $\latent$ and game parameters $\gameParams$---a modeling choice that is consistent with conventional \acp{vae}~\cite{kingma2013auto}.
Future work could extend our structured amortized inference pipeline to models that assume a \emph{stochastic} decoding process as in hierarchical \acp{vae}~\cite{sonderby2016ladder} or recent diffusion probabilistic models~\cite{ho2020denoising}.
Finally, our pipeline's flexibility allows for investigating different equilibrium concepts, such as entropic cost equilibria~\cite{mehr2023maxent}.

\begin{credits}
\subsubsection{\ackname}
We thank Yunhao Yang, Ruihan Zhao, Haimin Hu, and Hyeji Kim for their helpful discussions. This work was supported in part by the National Science Foundation (NSF) under Grants 2211432, 271643-874F, 2211548 and 2336840, and in part by the European Union under the ERC Grant INTERACT 101041863. Views and opinions expressed are those of the authors only and do not necessarily reflect those of the granting authorities; the granting authorities cannot be held responsible for them. 
\subsubsection{\discintname}
The authors have no competing interests to declare that are
relevant to the content of this article.
\end{credits}
\bibliographystyle{splncs04}
\bibliography{liu2023variational}

\end{document}